\documentclass[letterpaper, 10 pt, conference]{ieeeconf}  
\IEEEoverridecommandlockouts                              

\usepackage{amsmath,graphicx}
\usepackage[utf8]{inputenc} 
\usepackage[T1]{fontenc}    
\usepackage{hyperref}       
\usepackage{url}            
\usepackage{booktabs}       
\usepackage{amsfonts}       
\usepackage{nicefrac}       
\usepackage{microtype}      
\usepackage{graphicx}
\usepackage{booktabs}   
\usepackage{xcolor}

\renewcommand{\vec}[1]{\boldsymbol{#1}} 

\title{\LARGE \bf
A Two-Stage Multiple Instance Learning Framework for the Detection of Breast Cancer in Mammograms}

\author{Sarath Chandra K.$^{1}$, Arunava Chakravarty$^{1}$, Nirmalya Ghosh$^{1}$, Tandra Sarkar$^{2}$, \\Ramanathan Sethuraman$^{3}$,  Debdoot Sheet$^{1}$
\thanks{*This work is supported through a research grant from Intel India Grand Challenge 2016 for Project MIRIAD.}
\thanks{$^{1}$ S. Chandra, A. Chakravarty, N. Ghosh, T. Sarkar, R. Sethuraman, D. Sheet are with the Indian Institute of Technology Kharagpur, India-721302 ({\tt \{ sarathchandra.knv, arunava, nirmalya, debdoot\}@ee.iitkgp.ac.in})}%
\thanks{$^{2}$ T. Sarkar is with Apollo Gleneagles Hospital, Kolkata, India}%
\thanks{$^{3}$ R. Sethuraman is with Intel Technology India Pvt. Ltd. Bangalore, India }}

\begin{document}

\maketitle
\thispagestyle{empty}
\pagestyle{empty}

\begin{abstract}

Mammograms are commonly employed in the large scale screening of breast cancer which is primarily characterized by the presence of malignant masses. 
However, automated image-level detection of malignancy is a challenging task given the small size of the mass regions and difficulty in discriminating between malignant, benign mass and healthy dense fibro-glandular tissue. To address these issues, we explore a two-stage Multiple Instance Learning (MIL) framework. A Convolutional Neural Network (CNN) is trained in the first stage to extract local candidate patches in the mammograms that may contain either a benign or malignant mass. The second stage employs a MIL strategy for an image level benign vs. malignant classification. A global image-level feature is computed as a weighted average of patch-level features learned using a CNN. 
Our method performed well on the task of localization of masses with an average Precision/Recall of 0.76/0.80 and acheived an average AUC of 0.91 on the image-level classification task using a five-fold cross-validation on the INbreast dataset. Restricting the MIL only to the candidate patches extracted in Stage 1 led to a significant improvement in classification performance in comparison to a dense extraction of patches from the entire mammogram.

\end{abstract}

\section{INTRODUCTION}

Breast cancer is a leading cause of mortality in women \cite{desantis2017breast}. Its early detection through large-scale screening is critical for timely intervention but requires the manual evaluation of a large number of mammograms \cite{miller2014twenty}. However, only a small proportion of the entire screening population actually exhibits salient markers for malignancy. Thus, automated diagnostic tools can reduce the Radiologist's workload by referring only the suspicious cases to them and may also aid in minimizing the inter and intra-expert variations in diagnosis.

Breast cancer is primarily characterized by the presence of malignant lumps or tumors called mass in addition to other secondary indications such as the distribution of microcalcifications, asymetries,  and architectural distortions \cite{birads}. A non-cancerous benign mass typically has a smooth and regular shape while a malignant mass is usually characterized by irregular and indistinct margins. 
Many recent methods for screening breast cancer \cite{dhungel2017fully}, \cite{kooi2017large} first segment the lesions and then extract features from the detected lesions for image-level classification into a benign or malignant class.  In \cite{dhungel2017fully}, the lesion segmentation maps for multi-view images of the same breast were provided as input to an ensemble of ResNet models for breast level classification. In \cite{kooi2017large}, handcrafted features were augmented with Convolutional Neural Network (CNN) features to improve classification.  A transfer learning approach was employed in \cite{shen2019deep} where a patch level classifer was first trained to detect the lesions and then extended using additional convolutional layers at the end to obtain the image-level classifier. Alternatively, few single stage Multiple Instance Learning (MIL) approaches \cite{miccai_mil}  have been explored which extract features from dense patches obtained from the entire image without lesion segmentation. The sparse patch level prediction scores from a classifier are aggregated into image-level predictions using a max operation \cite{miccai_mil}.

\begin{figure}[]
 \centering
  \includegraphics[width=0.48\textwidth]{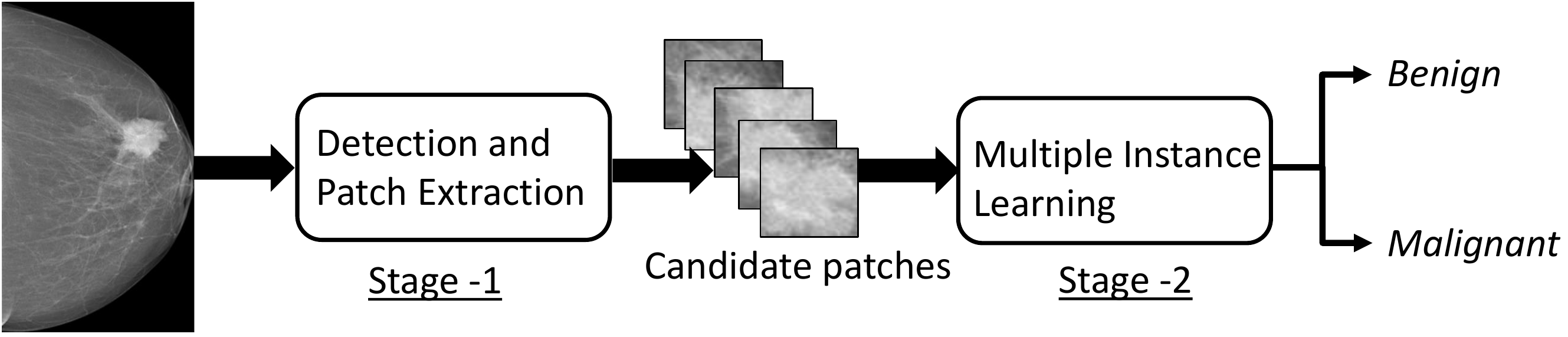}
 \caption{Pipeline of the proposed method.}
\label{Fig:pipeline} 
 \end{figure}
 
 \begin{figure*}[]
 \centering
  \includegraphics[width=0.9 \textwidth]{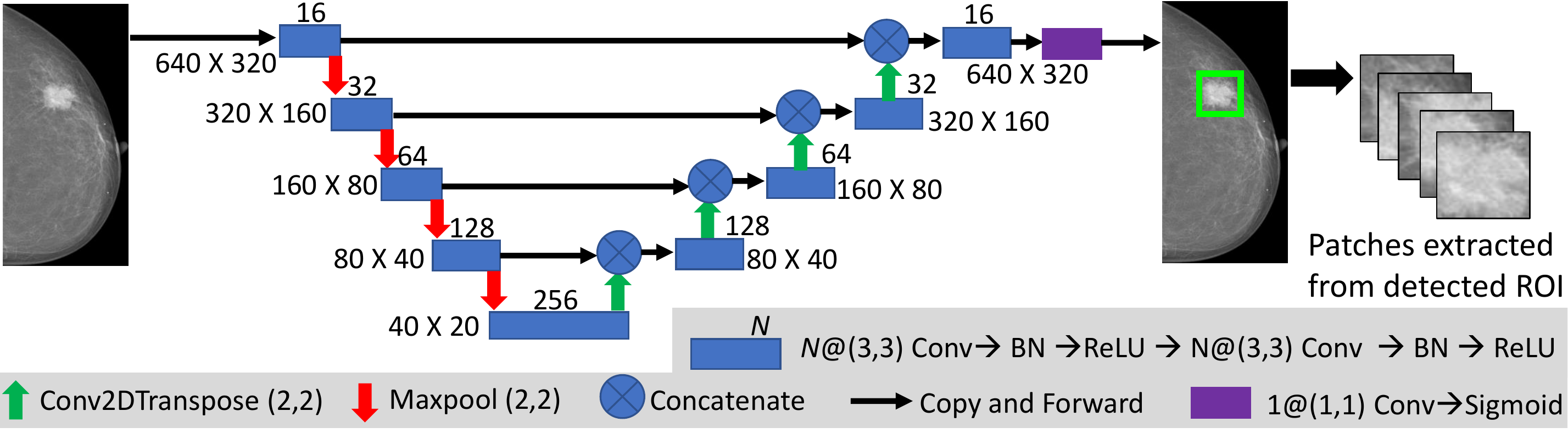}
 \caption{The block diagram and CNN architecture used in Stage 1 to detect the bounding boxes of the mass regions in the mammogram.}
\label{Fig:localization} 
 \end{figure*}

The image-level screening of mammograms is a challenging task. Since a mass occupies a very small region (around $2 \%$ \cite{miccai_mil}) in the entire mammogram, the classifier must learn to attend to localized suspicious regions while ignoring the large healthy background tissue. Moreover, discriminating between a benign, malignant mass and a dense healthy fibro-glandular tissue is a challenging task as they often share a similar intensity and texture profile. 
Finally, the explainability of an automated system in terms of an approximate localization of the mass which led to the prediction of cancer is critical for deployment in a real-world scenario.

To address these issues, we explore a two-stage framework depicted in Fig. \ref{Fig:pipeline}. A localization network is trained in the first stage to extract local candidate patches in the mammograms that may contain either a benign or malignant mass. In the second stage, a Multiple Instance Learning (MIL) formulation is employed to obtain a global image-level feature representation from the extracted image patches to classify the mammograms. In contrast to the existing single stage MIL methods, we hypothesize that restricting the MIL framework only to the regions detected in the first stage may make its task simpler resulting in improved classification performance. The proposed method has been found to provide a robust mass localization and competitive image-level classification performance on the INbreast dataset \cite{moreira2012inbreast}. The efficacy of utilizing the localization network in improving the image-level classification performance has also been demonstrated experimentally. The details of the proposed method is presented in Section \ref{Sec:Method} followed by a discussion of the experimental results in Section \ref{Sec:result} and concluding remarks in Section \ref{Sec:conclusion}.

\section{METHOD}
\label{Sec:Method}
The proposed method employs a two-stage framework (see Fig. \ref{Fig:pipeline}) to ensure that it only attends to the clinically relevant localized regions in the mammogram for its predictions. The first stage is a \textit{localization network} that detects multiple bounding boxes in each image which are the candidate regions in the mammogram that may contain either a benign or a malignant mass. A set of image patches are extracted from the detected regions and employed in the second stage for the image-level prediction. The second stage employs a MIL strategy to obtain a global image-level feature representation by computing a weighted average of  patch-level features. Both the aggregation weights and the patch-level features are obtained by applying a CNN model to each image patch. Further details are discussed below.

\noindent \textbf{Image Preprocessing: }
A tight crop around the breast region is obtained by roughly segmenting it using Otsu Thresholding to remove the dark background and resized to $640 \times 320$ pixels. The images are whitened to zero mean and unit variance. During training, the data augmentation is performed with random vertical and horizontal flips, translation and scaling within a range of 0.2 times the image dimensions, and rotations within $\pm 30 ^\circ$ to improve generalization. 

\noindent \textbf{Stage 1, Localization of Mass: } The localization network employs a Fully Convolutional Network similar to U-Net \cite{ronneberger2015u} with the number of convolutional filters in each layer modified for our task as depicted in Fig. \ref{Fig:localization}. The output of the localization network is a softmap where the value at each pixel is the probability ($\in [0,1]$) that it belongs to a mass. It is post-processed to obtain the detection bounding boxes as follows. First, the softmap is binarized by thresholding at $0.5$ and a morphological closing operating is applied to remove small spurious regions and smooth the boundaries of the detected regions. Thereafter, a tight bounding box is extracted around each of the detected connected components.

A linear combination of the weighted binary cross-entropy (WCE) loss (with a weight of 0.8) and the soft Dice loss \cite{milletari2016v} (with a weight of 0.2) is employed at a pixel-level to train the localization network. A weight of 28 is given to the positive class in the WCE loss based on the ratio of the number of foreground to background pixels in the Ground Truth (GT) bounding boxes. 

\noindent \textbf{Patch Extraction: }
Once the localization network has been trained, 5 candidate patches (one at the center and four from each corner) of size $64 \times 64$ pixels are extracted from each of the detected bounding boxes. We note that not all the extracted patches are malignant as some will contain benign masses while a few of them may be False Positives containing healthy fibro-glandular tissues that have a similar intensity and texture characteristics as that of a mass. In rare cases, where no masses were detected, a set of 5 patches were randomly selected from the bright regions in the entire image.

\begin{figure*}[]
 \centering
  \includegraphics[width=0.9 \textwidth]{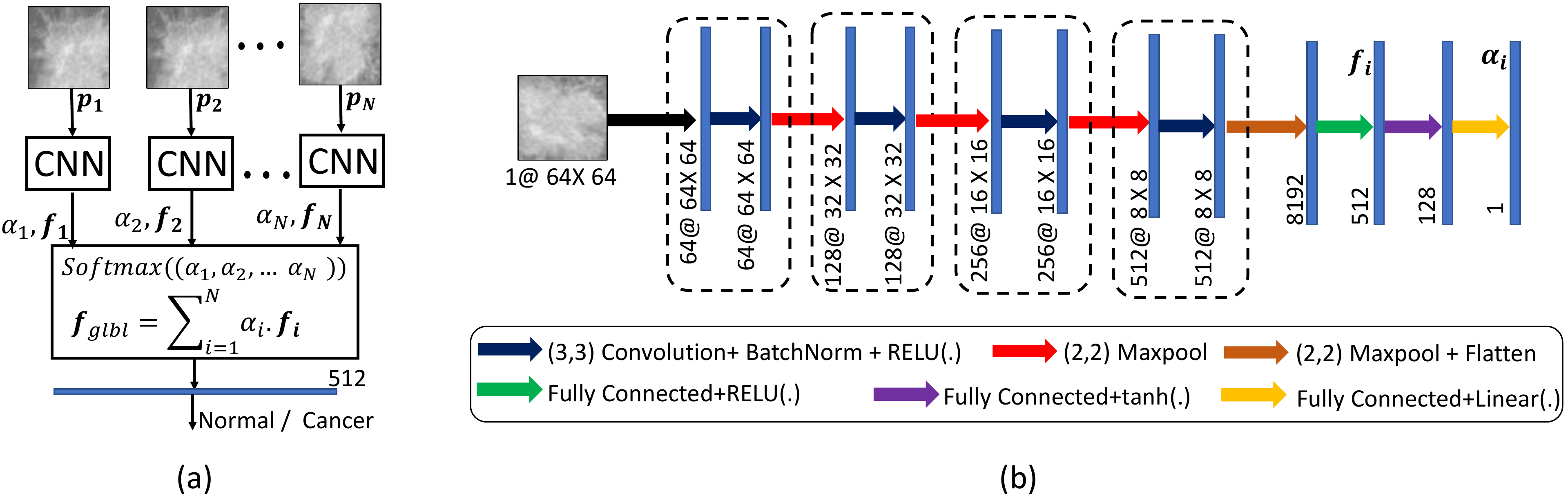}
 \caption{ The Multiple Instance Learning Framework in Fig. (a) extracts a set of local features $\vec{f_{i}}$ and corresponding scalar weights $\alpha _{i}$ by applying a single CNN model which shares the same model weights across the different patches. The image level feature $\vec{f}_{glbl}$ is obtained as the weighted average of the local features and used for benign vs. malignant classification using a FC layer. The architecture of the patch level CNN is detailed in Fig. (b). }
\label{Fig:mil} 
 \end{figure*}
 
\noindent \textbf{Stage 2, Image-level Classification:}
In Stage 2, our objective is to perform  a benign vs. malignant classification at the image-level by only attending to the set of $N$ candidate patches denoted  by  $\lbrace \vec{p}_{i}| 1 \le i \le N, i \in \mathbb{Z}^+ \rbrace$ which were extracted in the previous stage. $N$ may vary across different images. An image belongs to the positive class if \textit{at least one} $\vec{p}_{i}$ contains a malignant mass while \textit{none} of the $\lbrace \vec{p}_{i}| 1 \le i \le n \rbrace$ in a benign image can have a malignant mass. However, the GT class labels are only available at the image-level and not known for each individual patch. No specific ordering is assumed among the image patches. This image-level classification task can be posed as a MIL where each image is treated as a \textit{Bag} of $N$ \textit{Instances} which represent the image patches $\vec{p}_{i}$.

\textit{In contrast to the MIL approach in \cite{miccai_mil} which attempted to combine the individual predictions for each $\vec{p}_{i}$  using sparsity constraints and max-pooling operations, we directly obtain the image-level prediction without explicitly classifying each $\vec{p}_{i}$ individually.} We employ an embedding-level approach similar to \cite{ams_mil} which is depicted in Fig. \ref{Fig:mil} (a). Each image patch $\vec{p}_{i}$ is encoded  into a 512-dimensional feature vector $\vec{f}_{i}$ using a CNN model which additionally computes a scalar attention weight $\alpha _{i}$. Each $ \alpha _{i}$ is normalized using the $Softmax \left( \left ( \alpha _{1}, \alpha _{2}, ... \alpha _{N} \right ) \right ) $ operation. Finally, the global image-level feature is computed as $\vec{f}_{glbl}= \sum _{i=1}^{N} \alpha _{i} . \vec{f}_{i} $ and used for binary classification using an FC layer with 1 neuron and $Sigmoid()$ activation. The binary cross-entropy loss was used to train the entire MIL architecture in Fig. \ref{Fig:mil} (a) and the class imbalance was handled during training by oversampling the instances of the malignant class with data augmentation.

The $Softmax \left( \left ( \alpha _{1}, \alpha _{2}, ... \alpha _{N} \right ) \right ) $ operation normalizes the attention weights to sum to 1 thereby ensuring that the scale of $\vec{f}_{glbl}$ is invariant to $N$ and also induces a regional competition among the image patches. The same CNN model with identical network weights is applied to each $\vec{p}_{i}$ to obtain $\vec{f}_{i}$ and $\alpha _{i}$ whose architecture is detailed in Fig.  \ref{Fig:mil} (b).

\section{EXPERIMENTS}
\label{Sec:result}

\noindent \textbf{Dataset: } 
Similar to \cite{miccai_mil}, the proposed framework has been evaluated on the  INbreast dataset \cite{moreira2012inbreast}.
It consists of 410 Full-field Digital Mammographic images in both MLO and CC views from 115 subjects  out of which 310 are benign and 100 contain malignant masses. The Ground Truth (GT) contains a rough localization of the masses and the class labels for benign (0) or malignant (1) for each image. If a mammogram with cancer has multiple masses, at least one (or more) of the masses will be malignant but the GT for each mass is not available as class labels are only available at the image level. 

A stratified five-fold cross-validation was performed for evaluation. In contrast to \cite{miccai_mil} which partitioned the dataset at an image-level, we partitioned the five folds at the subject level thereby ensuring that the MLO/CC views of the same breast do not occur simultaneously in both the train and test splits in any fold. This ensures that the performance evaluation is not biased due to overfitting. The implementation of \cite{miccai_mil} made available by its authors\footnotemark is re-trained on our fold partitions for a fair benchmark comparison.

\noindent \textbf{Training Details:} 
The first and second stages are trained separately. Once the localization network has been trained, it is used to extract the image patches which are then used to train the second stage. The localization network in Stage 1 is trained  for 300 epochs with a batch size of 8 and 36 batch updates per epoch. The Stochastic Gradient Descent (SGD) optimizer is employed with a learning rate of 0.001, weight decay of 0.0005 and the learning rate is decayed by 0.1 after 50, 200 and 250 epochs. 

The MIL network in Stage 2 is trained for 100 epochs using the SGD optimizer with a learning rate of 0.001, weight decay of 0.0005 and the learning rate is decayed by 0.1 after 50 epochs. During implementation, each training batch is of a variable size $N$ and is composed of all image patches extracted from a single image.

\noindent \textbf{Implementation Details:} The proposed method is implemented in Python 3 using the PyTorch 1.0 library and trained on a server with $2$ Intel Xeon E5-2620 CPU with $64$ GB RAM, $2$ TB HDD and 3 Nvidia GTX TITAN X GPU with $12$ GB RAM.

\noindent \textbf{Performance of mass localization: } The objective of the localization network in Stage 1 is not an accurate semantic segmentation but to obtain an approximate localization of the masses in terms of bounding boxes to enable the extraction of the candidate image patches for the second stage. We also note that the MIL framework can learn to ignore the False Positives (FP) from the first stage by assigning $\vec{\alpha _{i}} \approx 0$ for these patches. A predicted bound box is treated as a True Positive (TP) if it has an Intersection over Union (IoU) greater than 0.5 with respect to the GT bounding box. The localization network performed reasonably well with an average \textit{Precision of $0.76 \pm 0.09$} and an average \textit{Recall of  $0.80 \pm 0.12$} across the five-fold cross-validation.

\footnotetext{github.com/wentaozhu/deep-mil-for-whole-mammogram-classification/} 

\begin{figure*}[]
 \centering
  \includegraphics[width=0.99\textwidth]{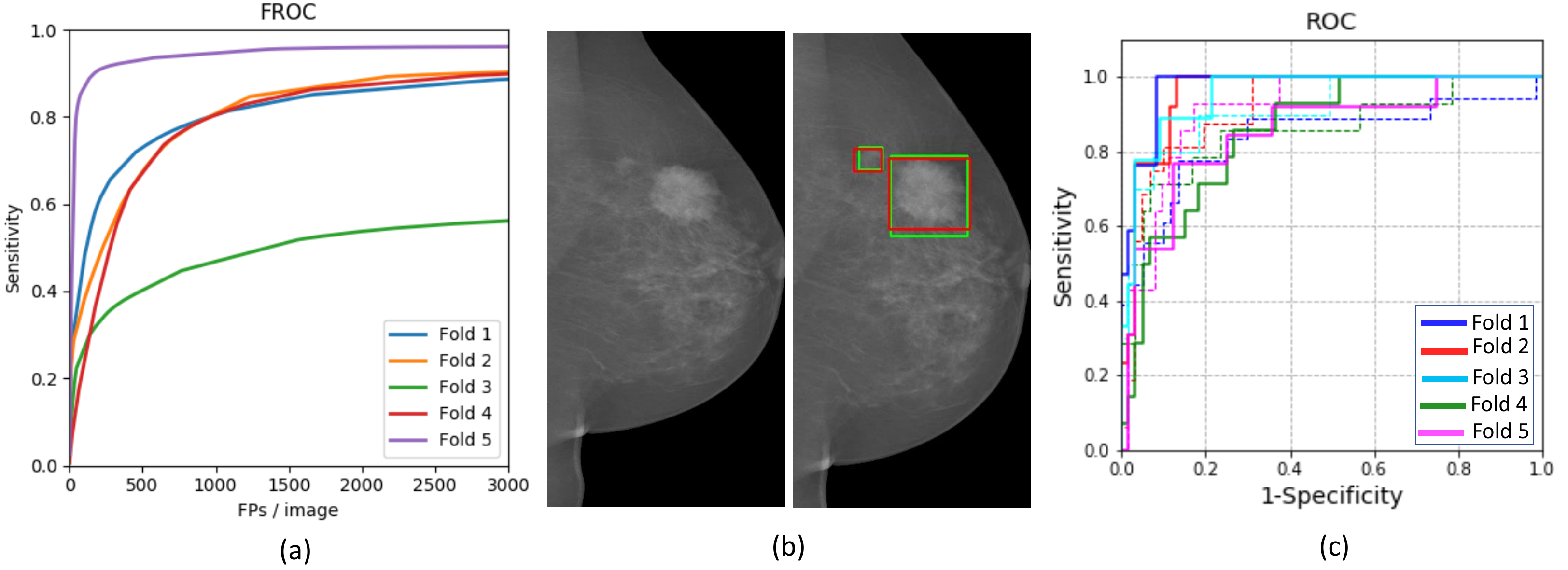}
 \caption{(a) FROC curves for the localization of mass in Stage 1 for the five folds. (b) A qualitative result of the mass localization in Stage 1 for the mammogram in left is depicted in the right image, with the GT marked in Red and our result in Green. (c) The ROC curves for the proposed image level benign vs. malignant classification for the five folds. The ROC curves for \cite{miccai_mil} is also plotted for each fold using dashed line of same color.}
\label{Fig:plots} 
 \end{figure*}

The localization performance has also been evaluated at a \textit{pixel} level using the FROC plots presented in Fig. \ref{Fig:plots} (a), where the TP (FP) is defined as the number of pixels in the predicted regions that lie within any (outside all) of the GT bounding boxes in each image. \footnote{Additional qualitative results for mass localization are available online at: \url{http://bit.do/EMBC2020_mammogram}}

\noindent \textbf{Performance of Image-level Classification: } The average image-level classification performance of the proposed method across the five folds and the corresponding ROC plot are presented in Table \ref{Tab_stage2} and Fig. \ref{Fig:plots} (c) respectively. 
The performance of employing the proposed MIL-framework alone without using the localization network in Stage 1 was also evaluated by extracting a dense set of overlapping image patches with a stride of $8$ pixels from the entire breast region and retraining our MIL framework. The proposed two-stage framework was found to significantly outperform Stage 2 alone (see Table \ref{Tab_stage2})  demonstrating the efficacy of the localization network in simplifying the classification task in the second stage by restricting it to the suspicious regions in the image.


Our method with an average AUC of $0.91$ outperforms the existing MIL based method in \cite{miccai_mil} by $\frac{0.91-0.89}{0.89} \times 100=2.25 \%$ in terms of AUC. Moreover, our method has a \textit{significantly better Sensitivity-Specificity trade-off in comparison to \cite{miccai_mil}} (see Table \ref{Tab_stage2}) which is desirable in screening systems.

\renewcommand{\arraystretch}{1.2}
\begin{table}[]
\centering
\caption{\scriptsize{Average five-fold cross-validation performance for image-level Benign vs. Malignant Clasification. The Sensitivity (Sens.), Specificity (Spec.), Balanced Accuracy (B. Acc.) and Area under ROC curve (AUC) are reported as mean $\pm$ standard deviation.}}
\label{Tab_stage2}
\resizebox{0.48 \textwidth}{!}{
\begin{tabular}{@{} lcccc @{}}
\toprule
                       & Sens. & Spec. & B. Acc. & AUC \\ \midrule
Stage 2 alone &  $0.84 \pm 0.04$    &   $0.36 \pm 0.06$   &    $0.60 \pm 0.10 $     &    $0.70 \pm 0.09$ \\ 
Proposed               & $0.96 \pm 0.04$     & $0.77 \pm 0.13 $     &   $0.86 \pm 0.08$     &  $0.91 \pm 0.06  $  \\
\cite{miccai_mil}  &  $ 0.90 \pm 0.02 $     & $ 0.72 \pm 0.17$    & $ 0.81 \pm 0.08$             &  $0.89 \pm 0.05$   \\ \bottomrule
\end{tabular}
}
\end{table}

\section{CONCLUSION}
\label{Sec:conclusion}

Mammograms are commonly employed in the large scale screening of breast cancer which is primarily characterized by the presence of malignant masses. However, image-level detection of cancer poses many challenges due to the small size of the mass regions in the entire image and the difficulty in discriminating between malignant, benign mass and healthy dense fibro-glandular tissues. 

In this work, we propose a two-stage framework to address these issues. In the first stage, bounding boxes around the mass regions are detected which are used to extract a set of candidate image patches. The second stage  employs a MIL strategy to obtain a global image-level feature representation by computing a weighted average of patch-level features learned using a CNN. Our method performed reasonably well on the task of localization of masses with an average Precision/Recall of $0.76/0.80$ and achieved an average AUC of 0.91 on the image-level classification task outperforming a state-of-the-art MIL based method.
Finally, performing a more fine-grained 6-level classification in the second stage to predict the BIRADS severity scale \cite{birads} presents a challenging direction for future work.

\bibliographystyle{unsrt}
\bibliography{root}
\onecolumn
\setcounter{page}{1}

\section*{Supplementary Material: A Two-Stage Multiple Instance Learning Framework for the Detection of Breast Cancer in Mammograms}
\vspace{-10 pt}

 \begin{figure*}[h]
 \centering
  \includegraphics[width=.66 \textwidth]{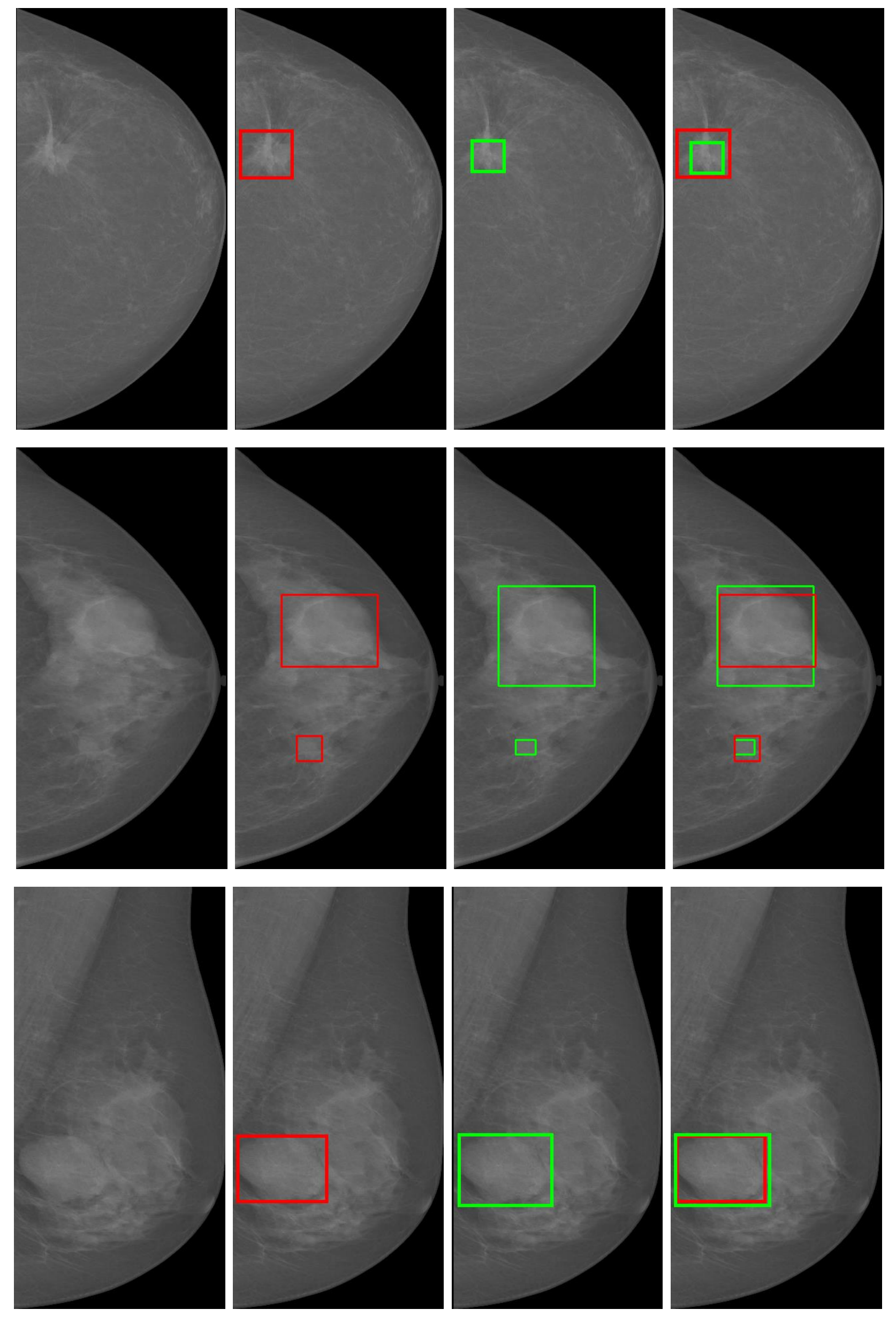}
 \caption{The qualitative results of the mass detection using the localization network in Stage 1. First column: the input image; Second Column: The Ground Truth(GT) bounding box in Red; Third Column: The predicted Bounding box in Green; Fourth Column: The GT and predicted detections are overlayed.}
\label{Fig:supplementary} 
 \end{figure*}

\end{document}